\documentclass[10pt,twocolumn,letterpaper]{article}

\usepackage[pagenumbers]{cvpr} 

\usepackage{multirow}
\usepackage{pifont}
\usepackage{todonotes}
\usepackage{tabularx}

\newcommand{\cmark}{\ding{51}}%
\newcommand{\xmark}{\ding{55}}
\newcommand{\checkmarkgreen}{\textcolor{my_green}{\cmark}}
\newcommand{\xmarkred}{\textcolor{red}{\xmark}}
\definecolor{my_green}{HTML}{339933}
\definecolor{my_red}{HTML}{eb9898}
\definecolor{my_green2}{HTML}{e1f5e6}

\definecolor{cvprblue}{rgb}{0.21,0.49,0.74}
\usepackage[pagebackref,breaklinks,colorlinks,allcolors=cvprblue]{hyperref}
\newcommand{\mypar}[1]{\vspace{2mm}\noindent\textbf{#1}}

\newcommand{\ab}{\mathbf{a}}
\newcommand{\vb}{\mathbf{v}}

\newcommand{\ours}{AVH-Align\xspace}
\newcommand{\ourssup}{AVH-Align/sup\xspace}

\title{Circumventing shortcuts in audio-visual deepfake detection datasets with unsupervised learning}

\author{
Stefan Smeu\thanks{Equal contribution.} $^{1}$ ~~~
Dragos-Alexandru Boldisor\footnotemark[1] $^{1}$ ~~~
Dan Oneata$^2$ ~~~
Elisabeta Oneata$^1$ \\
$^1$Bitdefender ~~~~~
$^2$\textsc{Politehnica} Bucharest \\
{\tt\small\{ssmeu, dboldisor, eoneata\}@bitdefender.com} ~~
{\tt\small dan.oneata@gmail.com}
}

\begin{document}
\maketitle

\begin{abstract}
Good datasets are essential for developing and benchmarking any machine learning system. Their importance is even more extreme for safety critical applications such as deepfake detection---the focus of this paper. Here we reveal that two of the most widely used audio-video deepfake datasets suffer from a previously unidentified spurious feature: the leading silence. Fake videos start with a very brief moment of silence and, on the basis of this feature alone, we can separate the real and fake samples almost perfectly. As such, previous audio-only and audio-video models exploit the presence of silence in the fake videos and consequently perform worse when the leading silence is removed. To circumvent latching on such an unwanted artifact and possibly other unrevealed ones, we propose a shift from supervised to unsupervised learning by training models exclusively on real data. We show that by aligning self-supervised audio-video representations we remove the risk of relying on dataset-specific biases and improve robustness in deepfake detection.

\end{abstract}
\section{Introduction}


Manipulated videos represent a threat to society as they have the potential of misleading people into believing actors with malicious intents.
By spreading misinformation on social media platforms, people may be exposed to scams (e.g., identity theft operations), conspiracy theories and political misinformation.
Therefore, deepfake detection methods are essential tools on a global scale.


The progress in automated deepfake detection is fueled
by the datasets 
developed by the research community. 
Good quality data is essential for both training and benchmarking the progress of these methods.
In recent years, numerous datasets have been proposed.
Among the 
audio-video deepfake datasets,
there have been released datasets that alter both streams \cite{khalid2022fakeavcelebnovelaudiovideomultimodal, cai2024avdeepfake1mlargescalellmdrivenaudiovisual}
or only one \cite{wang2020asvspoof2019largescalepublic, muller2024mlaad, Rossler_2019_ICCV} ; datasets
 with full-video  \cite{khalid2022fakeavcelebnovelaudiovideomultimodal} or only local   \cite{cai2024avdeepfake1mlargescalellmdrivenaudiovisual, cai2023glitch} manipulations.

However, care has to be taken to ensure a good deepfake detection dataset.
Any asymmetry in the preparation of fakes and reals can result in 
biases that correlate spuriously with the groundtruth label.
For example, in the image domain, different preprocessing pipelines \cite{chai2020eccv} or
types of resizing \cite{rajan2024} between real and fake samples 
paint an overly optimistic picture.
In the audio domain, the very popular ASVSpoof19 dataset \cite{wang2020asvspoof2019largescalepublic} has been shown to leak information about labels in the form of silence duration \cite{muller2021speech} or bitrate information \cite{borzi2022synthetic}.
Since many of the deepfake detection models are high capacity,
they can easily 
use such %
artifacts as shortcuts for learning,
consequently greatly impacting their 
ability to generalize in a real case scenario.

\begin{figure}[t!]
    \centering
    \includegraphics[width=0.85\linewidth]{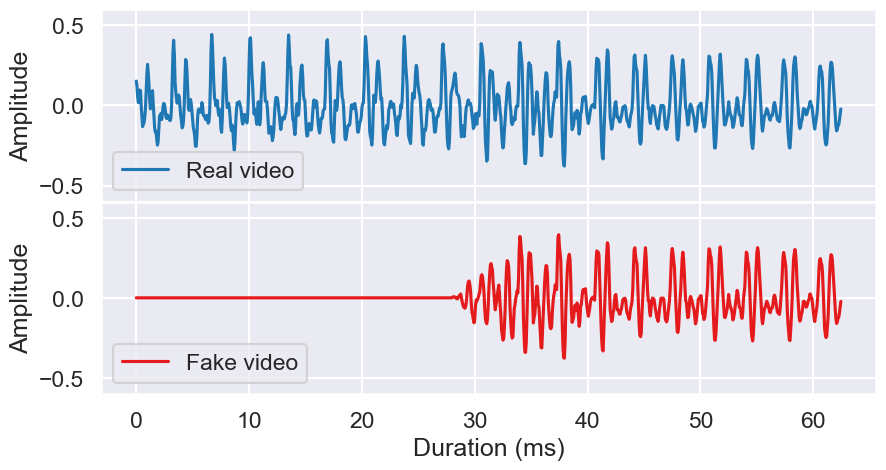}
    \caption{
    Audio-visual deepfake detection datasets have a silence bias:
    fake samples start with a brief moment of silence,
    which is not the case for real samples.
    Here we show the first 62.5 ms of the audio waveform for a real and the corresponding fake sample from the AV-Deepfake1M dataset \cite{cai2024avdeepfake1mlargescalellmdrivenaudiovisual}.
    }
    \label{fig:real_vs_fake}
\end{figure}

In this work, we first expose 
a bias present in two widely-adopted audio-video deepfake detection datasets,
FakeAVCeleb \cite{khalid2022fakeavcelebnovelaudiovideomultimodal} and
AV-Deepfake1M \cite{cai2024avdeepfake1mlargescalellmdrivenaudiovisual}:
a short moment of silence present at the beginning
of manipulated videos (see Figure~\ref{fig:real_vs_fake}).
Based on this information, a very simple silence classifier can reach a near-perfect performance of over 98\% on both FakeAVCeleb and AV-Deepfake1M.
Moreover, we observe that the silence distribution is similar across datasets,
implying that prior methods may have potentially over estimated the generalization performance based on this shortcut. 

Introducing
such 
biases
when creating the datasets is unavoidable and there might be others that are more subtle and harder to reveal.
To circumvent this problem, we show that one 
solution is a shift in learning paradigm, from the more common supervised setup to the unsupervised one \cite{feng2023self,reiss2023detectingdeepfakesseeing}.
Removing fake samples from training eliminates the focus 
on the asymmetries induced by spurious artifacts.
Moreover, this also has the potential to improve generalization among different manipulation techniques,
since supervised detection methods tend to over-rely on generator-specific fingerprints 
\cite{marra2019gans, yu2019attributing}.
By limiting 
to real data, we can also naturally leverage self-supervised representations, which have been shown to improve generalization \cite{ojha2023towards}.


To this end, we propose an approach, named \ours,
that learns a frame-level audio--video alignment score on top of AV-HuBERT features \cite{shi2022learning}.
Since tampered videos are expected to have greater audio-video desynchronizations, this alignment score can effectively differentiate fake and real videos. 
We show that this approach is robust with respect to the identified shortcut of leading silence,
and also outperforms other methods that do not use the silence bias, while not seeing any fake samples at training time.

To summarize, our work makes the following contributions:
1. We expose a previously unknown spurious feature in two of the most widely adopted deepfake detection datasets.
2. We analyze the impact of this shortcut on various state-of-the-art models.
3. We show that a way to mitigate such shortcuts is by training on real data only and we introduce a new method in this direction.
Our code is available at: {\small\url{https://github.com/bit-ml/AVH-Align}}.




\section{Related work}

\mypar{Audio-video deepfake detection.}
Many approaches for deepfake detection on videos have focused on the visual stream of information only
\cite{Afchar_2018,haliassos2021lipsdontliegeneralisable,FTCNiccv21,haliassos2022leveraging,implicitcvpr23,shiohara2022detectingdeepfakesselfblendedimages,cai2023marlinmaskedautoencoderfacial}.
But with the recent introduction of audio-visual datasets,
(FakeAVCeleb~\cite{khalid2022fakeavcelebnovelaudiovideomultimodal},
Deepfake TIMIT~\cite{korshunov2018deepfakesnewthreatface,salvi2023access},
KODF~\cite{Kwon_2021_ICCV},
LAV-DF~\cite{cai2023glitch},
AV-Deepfake-1M~\cite{cai2024avdeepfake1mlargescalellmdrivenaudiovisual}),
more research has shifted towards models that exploit both audio and video cues \cite{feng2023self,oorloff2024avff,shahzad2023avlipsyncleveragingavhubertexploit,hashmi2023avtenetaudiovisualtransformerbasedensemble,avoiddf2023,koutlis2024dimodiff}.
An emerging trend in this direction is the use of pretrained representations in a self-supervised way \cite{feng2023self,oorloff2024avff}.
But, different from our approach,
these methods train the representations from scratch and use them as a first step in a more elaborated pipeline:
supervised classification \cite{oorloff2024avff} or anomaly detection \cite{feng2023self}.
There are also works that similarly to us exploit the pretrained audio-video AV-HuBERT model \cite{shi2022learning} to extract representations \cite{shahzad2023avlipsyncleveragingavhubertexploit,hashmi2023avtenetaudiovisualtransformerbasedensemble}, but all of those methods are trained in the fully supervised learning paradigm.


\mypar{Unsupervised deepfake detection.}
To make detection more generalizable across generators,
a new direction 
is to 
depart from the supervised 
paradigm and resort only to real data.
There are two main classes of such unsupervised approaches:
methods that rely on consistency checks and
methods that treat the problem as an anomaly detection one.
Among the first class,
prior work proposed to verify that audio and visual streams align at a semantic level, for example, from the point of view of spoken content \cite{li2024zero,bohacek2024lost},
or at a representation level, based on the alignment of audio and video features \cite{feng2023self,reiss2023detectingdeepfakesseeing}.
Consistency checks have also been used with respect to the identity of a speaker (comparing a query sample against real audio \cite{Pianese_2024} or real images \cite{reiss2023detectingdeepfakesseeing} of the target speaker) or between image and text modalities \cite{reiss2023detectingdeepfakesseeing}.
For the second class, deepfake detection as anomaly detection,
recent work in the image domain has used the reconstruction loss of the query image to tell whether it is anomalous \cite{ricker2024cvpr,cozzolino2024eccv};
for example, Ricker \etal \cite{ricker2024cvpr} make the observation that images generated by a latent diffusion model (LDM) are easier to reconstruct by the LDM than real images.
For the audio-visual domain, Feng \etal \cite{feng2023self} use both classes of approaches:
they use consistency checks to estimate synchronization between the two streams and
then flag anomalies using density estimation.

\section{Silence bias in audio-video datasets}
\label{sec:silence-bias}

In this section,
we show that two popular datasets (Sec.~\ref{subsec:datasets}) have a silence bias.
We analyze its behavior and show that a simple classifier based on the leading silence alone can obtain almost perfect separation between fake and real samples (Sec.~\ref{subsec:analysis-of-leading-silence}).
This implies that the performance of prior work is susceptible to have been overestimated.
For this reason we analyze its impact on various audio and audio-visual methods (Sec.~\ref{subsec:impact-on-prior-work}).

\subsection{Datasets}
\label{subsec:datasets}

We consider two audio-visual datasets in our analysis:
FakeAVCeleb \cite{khalid2022fakeavcelebnovelaudiovideomultimodal} and AV-Deepfake1M \cite{cai2024avdeepfake1mlargescalellmdrivenaudiovisual}. 
They distinguish mainly in the fact that
the first contains fully-generated video sequences,
while the second contains partially-manipulated sequences.
Both are based on the VoxCeleb2 dataset \cite{Chung18b},
which consists of YouTube audio-video of celebrities.
Apart from the real samples%
---real video real audio (RVRA)---%
both datasets include three types of fake videos:
real video fake audio (RVFA),
fake video real audio (FVRA), and
fake video fake audio (FVFA).

\mypar{FakeAVCeleb} 
contains 500 real videos from VoxCeleb2
and 19.5k fake videos (10k FVFA, 9k FVRA, 500 RVFA).
The fake visual content was generated with  
face swapping methods (Faceswap \cite{korshunova2017fastfaceswapusingconvolutional} and FSGAN \cite{nirkin2019fsgansubjectagnosticface})
or the Wav2Lip lip syncing approach \cite{Prajwal_2020}.
The fake audio content was generate with the voice cloning tool SV2TTS \cite{jia2019transferlearningspeakerverification}.
The dataset is diverse across age groups, genders, races,
as well as with respect to the number of subjects in a single video, their placement, the visual and audio quality.
In our experiments, we split the dataset in 70\% (train and validation) and 30\% (test).
This split is kept consistent across all experiments.

\mypar{AV-Deepfake1M} is a large scale dataset,
which consists of over one million videos and 2k subjects.
As opposed to the FakeAVCeleb dataset, the manipulations here are local and consists of
word-level replacements, insertions and deletions.
The text manipulations are generated with the ChatGPT large language model.
The fake video content is generated with the lip syncing method TalkLip \cite{wang2023seeingsaidtalkingface},
while the fake audio content is generated with the VITS \cite{kim2021conditionalvariationalautoencoderadversarial} or YourTTS \cite{casanova22icml}
methods.
The authors ensure that the synthesized words share the same background noise with the full audio, by first extracting the audio noise with the Denoiser method \cite{defossez2020realtimespeechenhancement} and then adding it to the synthesized words.
The dataset is originally split into train, validation and test, with the test split having a different set of speakers than those encountered in the train and validation splits.
For our experiments, we select training and validation samples from the original training split,
and evaluate on 10k samples from the original validation split or on the full official test set.



\begin{figure}
    \centering
    \includegraphics[width=1\linewidth]{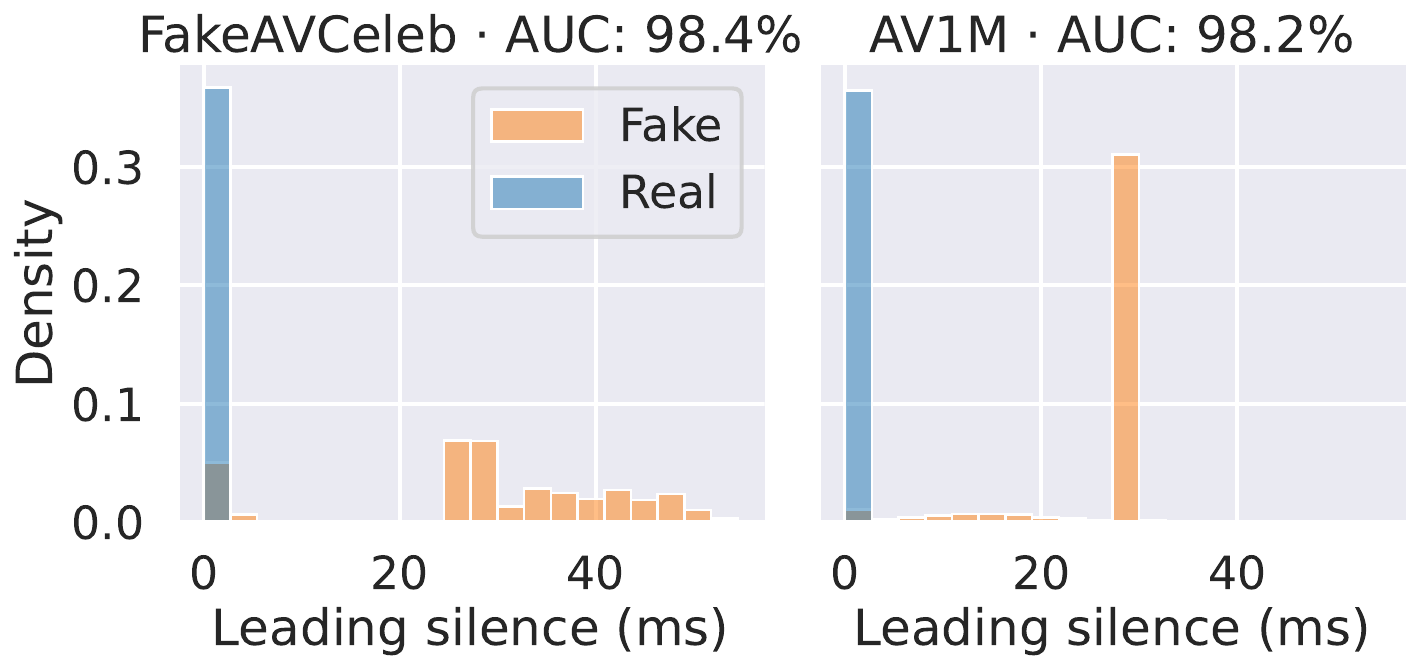}
    \caption{%
        Normalized distribution plots of the leading silence duration for real and fake videos in the
        FakeAVCeleb (left) and AV-Deepfake1M (right) datasets.
        The fake samples start with 25--30 ms of silence.
    }
    \label{fig:distribution_starting_points}
\end{figure}

\subsection{Analysis of leading silence}
\label{subsec:analysis-of-leading-silence}

We start by analyzing the silence distribution of the real and fake samples in the two considered datasets. 
We define the duration of the leading silence as the moment when the magnitude of the audio exceeds a certain threshold~$\tau$.
For this experiment, we select this threshold to be $5 \cdot 10^{-4}$, but as we will shortly see, the results are robust to its choice.
We carry this analysis only on the real (RVRA) and fully fake (FVFA) videos from each dataset's test set.

The results are shown in Figure \ref{fig:distribution_starting_points}.
We observe that the real videos start with noise,
while the fake samples have a leading silence of around 25--30 ms.
The silence duration of fake samples is similar for both datasets,
although the distribution is much sharper for the AV-Deepfake1M dataset.
If we were to rank the samples based on this feature we would obtain an area under curve 
of the receiver operating characteristic curve (AUC) of over 98\% for either datasets.

\begin{figure}
    \centering
    \includegraphics[width=1\linewidth,trim={{0.8cm} {0.8cm} {0.6cm} {0.8cm}},clip]{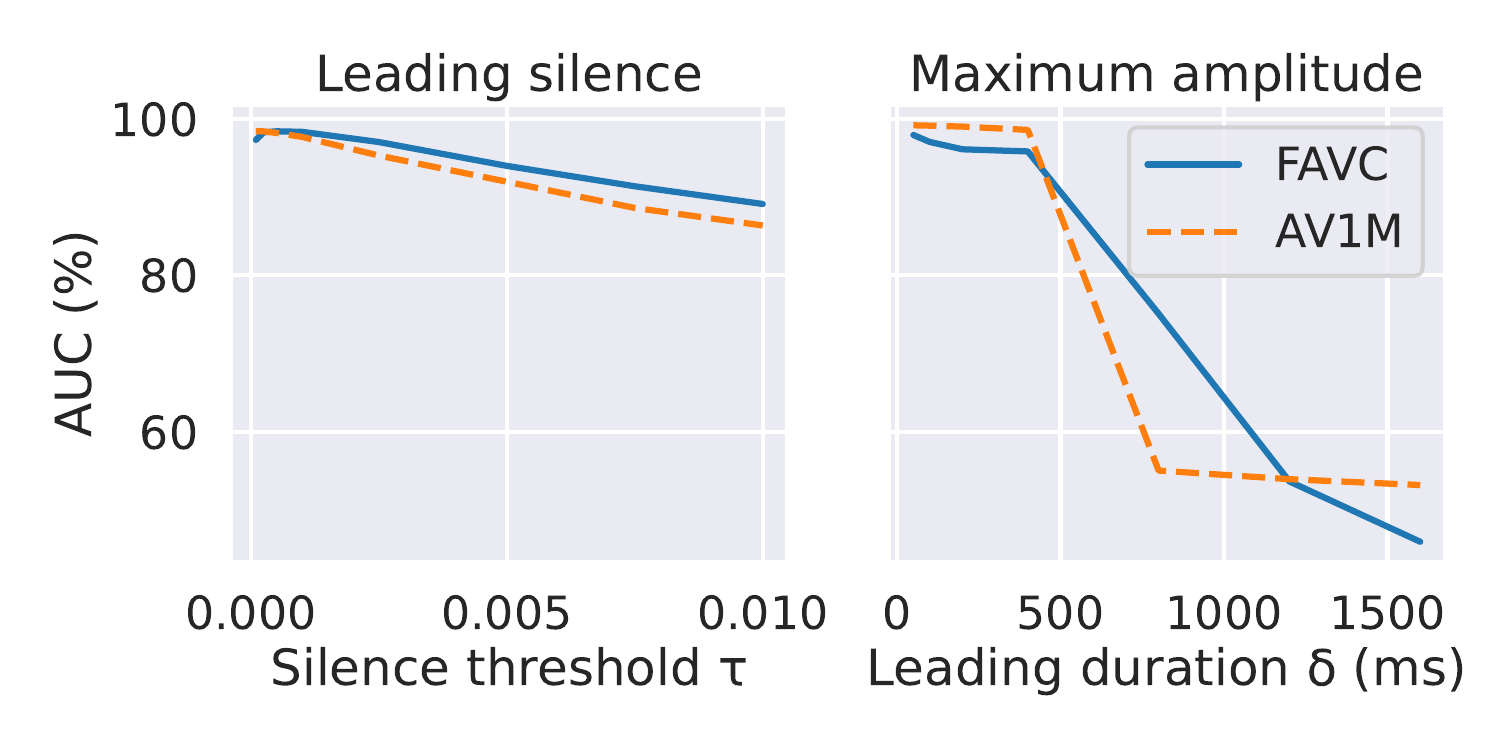}
    \caption{%
        Left: The impact of the silence threshold on the leading silence classifier.
        Right: The impact of the leading duration on the maximum amplitude classifier.
    }
    \label{fig:impact-of-params}
\end{figure}

\mypar{What counts as silence?}
For the previous experiment we have considered that silence is the signal that has an amplitude lower than $\tau = 5 \cdot 10^{-4}$.
We investigate how sensitive the performance is to this threshold.
We vary $\tau$ across a grid of values and
show the results in Figure \ref{fig:impact-of-params} (left).
We observe that the results are strong as long as this threshold is small enough.

\mypar{A different perspective: Maximum amplitude.}
Instead of looking at the silence duration,
we can alternatively measure the maximum amplitude in the first $\delta$ seconds of the audio.
Figure~\ref{fig:impact-of-params} shows the AUC obtained by ranking the samples based on this feature for various leading durations $\delta$.
We observe that this alternative measure also yields strong results (over 98\% AUC),
and that the optimum is obtained for a $\delta$ of around 30 ms, in line with the previous experiment.
As we extend the window, the performance gets closer to random chance, AUC of 50\%.

\mypar{Other biases: Volume and trailing silence.}
If we extend the duration $\delta$ to cover the entire audio,
we get an estimate of the maximum volume of the audio.
This feature yields an AUC of 67.6\% on FakeAVCeleb,
which is much less than 98.4\%, but still over random chance (50\% AUC).
Similarly, we have also investigated the trailing silence (the silence at the end of the audio) and have obtained AUC of over 99\% for FakeAVCeleb. 
AV-Deepfake1M is less sensitive to these other biases, with values closer to 50\%.
While these biases are not as consistent as the leading silence, they are still problematic.

\mypar{Why do fake samples have a leading silence?}
Given that we do not have access to generation process of the two datasets,
it is challenging to pinpoint the exact reason for the occurrence of the leading silence bias.
However, we speculate that this happens when the audio may be slightly shorter than the video counterpart.
Note that this is a different reason from the silence observed in audio only datasets \cite{muller2021speech}.
There the real audios had a leading silence, while the synthesized speech was silence-free.
In the case of audio-video datasets it might be challenging to completely avoid this problem,
but an easy solution (for this particular bias) is to trim the leading silence.
This is what we do in the next experiments.

\mypar{Biases in other datasets.}
We consider three more datasets: LAV-DF \cite{cai2023glitch}, AVLips \cite{liu2024lips}, DFDC \cite{dolhansky2020deepfake}.
We find that LAV-DF has a slight trailing silence bias (65\% AUC),
but it does not exhibit the other two biases (leading silence and volume).
In the case of AVLips and DFDC we did not identify any of the three biases. 
This may happen because the samples in these datasets have real audio and the unmodified audio matches the visual sequence more precisely.
This is confirmed by a fine-grained analysis on FakeAVCeleb,
where we observe that on the FVRA split the biases are diminished (62.4\% AUC for lead silence),
while they persist on the RVFA split (100\% AUC for lead silence).

\subsection{Impact on prior work}
\label{subsec:impact-on-prior-work}

We have shown that real and fake samples differ in terms of the leading silence.
This is a simple feature which could be learned by the high-capacity neural networks.
Here we investigate whether that is indeed the case for three existing methods:
\begin{itemize}
    \item \textbf{RawNet2} \cite{9414234}, which is
    an audio-only method that operates on the raw waveform.
    Its architecture sequences sinc layers, convolutions, gated recurrent units and fully connected layers.
    \item \textbf{MDS} \cite{chugh2020not} (modality dissonance score), which is
    an audio-visual method that estimates the mismatch between audio and video segments.
    The score is computed as the distance between audio and visual features.
    \item \textbf{AVAD} \cite{feng2023self} (audio-visual anomaly detection), which is
    an audio-visual method trained on real data only.
    The approach has two steps:
    first, it estimates the desynchronization between audio and video; 
    then it estimates whether these patterns are typical of real data or anomalous. 
\end{itemize}

We train the RawNet2 and MDS methods on both datasets using the code provided by the authors.
For the AVAD we use the provided checkpoint (trained on LRS \cite{son2017lip}) and do not retrain it since training code is not available.
As a baseline we include our leading silence duration classifier described in the previous section.

We evaluate all methods in two settings:
on the corresponding evaluation set and
on a trimmed version of the same evaluation set.
For the trimmed version, we discard the beginning of videos such that each of them starts with the first non-silent segment. We find the duration to discard by first computing the leading silence (using the same $5\cdot10^{-4}$ threshold on the magnitude) and then rounding it up to the nearest multiple of 1 / FPS (the reciprocal of the video's frame rate);
this rounding ensures that the audio and video channels remain synchronized. 
For training, we do not perform trimming, but use the original dataset.

The results are shown in Table~\ref{tab:fakeavceleb_trimmed}.
First, we observe that the audio-based RawNet2 is the top performing method.
Moreover, we see that the leading silence classifier is generally better than the two other approaches: MDS and AVAD. 
The relatively stable performance of AVAD indicates that this method dose not latch on the silence information.

In terms of the impact of the leading silence,
we observe that removing it through trimming affects methods differently.
As expected, the silence classifier is affected the most and
its performance drops down to random chance on the trimmed data.
RawNet2 is not as affected on the FakeAVCeleb,
presumably because there is still enough information throughout the signal,
but it suffers a larger hit on the more challenging AV-DeepFake1M,
which has only partially-manipulated samples. MDS is affected by the leading silence bias on both FakeAVCeleb and AV-Deepfake1M, though significantly stronger on AV-Deepfake1M. 
AVAD is the most robust method,
since this approach does not explicitly model the silence information.
Finally, we notice that the RawNet2 audio-only method is best even on the trimmed settings.
This is noteworthy since it is the first time an audio-only method has been applied to these datasets.
This suggests that the audio stream is an important source of information that is often overlooked by prior work.
\newcommand{\trr}[1]{\tiny{\textcolor{red}{$\downdownarrows$ #1}}}
\newcommand{\trre}[1]{\tiny{\textcolor{orange}{$\downarrow$ #1}}}
\newcommand{\trg}[1]{\tiny{\textcolor{my_green}{$\cong$ #1}}}

\begin{table}
    \centering
    \footnotesize
    \tabcolsep 4.1pt
    \begin{tabular}{%
            lc
            rr@{\hskip 0.05in}l
            rr@{\hskip 0.05in}l
        }
        \toprule
        &  & \multicolumn{3}{c}{FakeAVCeleb}  & \multicolumn{3}{c}{AV-Deepfake1M} \\
        \cmidrule(lr){3-5} \cmidrule(lr){6-8}
        Method  & Mod.
        & \multicolumn{1}{c}{Trim: \xmarkred} & \multicolumn{2}{c}{Trim: \checkmarkgreen}
        & \multicolumn{1}{c}{Trim: \xmarkred} & \multicolumn{2}{c}{Trim: \checkmarkgreen}
        \\
        \midrule
        Silence classifier      & A  &  98.4 & 54.8 & \trr{43.6} &  98.2 & 50.6 & \trr{47.6} \\
        RawNet2 \cite{9414234}   & A  & 99.9 & 97.3 & \trre{2.6}  & 99.9 & 88.1 & \trr{11.8} \\
        MDS \cite{chugh2020not}  & AV &  90.4 & 73.8 & \trr{16.6} &  99.2 & 54.9 & \trr{44.3} \\
        AVAD \cite{feng2023self} & AV &  95.2 & 95.2 & \trg{0.0}  &  52.9 & 52.9 &  \trg{0.0} \\
        \bottomrule
    \end{tabular}
    \caption{%
        The impact of leading silence on the performance (area under the receiver operator characteristic curve; AUC) for three existing deepfake detection methods and the silence classifier.
        Results are shown for RVRA-FVFA subsets.
    }
    \label{tab:fakeavceleb_trimmed}
\end{table}



\section{Modeling real data for deepfake detection}
\label{sec:methodology}






The previous section indicated that audio-visual datasets exhibit a silence bias which can easily be exploited.
We want to develop a method that is robust to this bias (and possibly other uncovered ones) and still performs well.
We have seen that models trained only on real data \cite{feng2023self} are promising in being robust to the silence shortcut, but the performance was modest.
To further improve them we propose to build on top of audio-focused self-supervised features. Self-supervised features have shown strong generalization for both visual \cite{ojha2023towards, Cozzolino_2024_CVPR} and audio \cite{pascu2024towards, Pianese_2024} deepfake detection.
We choose audio-focused self-supervised features because the audio models showed strong performance in the previous section.
Note that we cannot rely on audio-only models because there are cases where manipulations appear only in the visual domain (the fake video, real audio case).

\begin{figure*}[htb!]
   \centering
   \includegraphics[height=4cm]{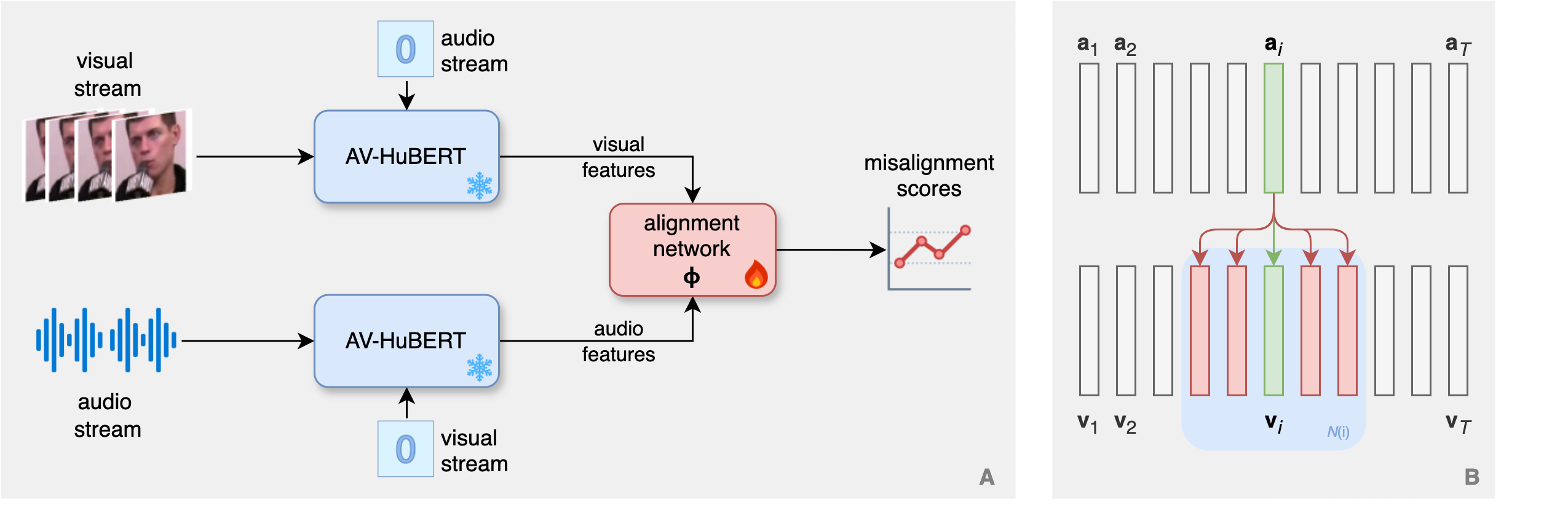}
   \caption{
    Overview of the \ours method.
    \textbf{A:} 
    We use the pretrained AV-HuBERT model to extract self-supervised features which we further align with a learnable network $\Phi$.
    Note that we use a single AV-HuBERT model, but make two forward passes to obtain audio-only and video-only features (instead of a single set of multimodal features).
    \textbf{B:} At training we maximize the alignment score $\Phi_{ii}$,
    between the audio features $\ab_i$ at time step~$i$ and the corresponding video features $\vb_i$,
    while minimizing the alignment $\Phi_{ik}$ to the other features $\vb_k$ in a neighboring window $\mathcal{N}(i)$.
    }
   \label{fig:method}
\end{figure*}

\subsection{Method}

We propose a method that aligns AV-HuBERT \cite{shi2022learning} features on real data.
First, we extract audio and visual frame-level features with a pretrained AV-HuBERT model.
Then, on top of these features we learn a network to better align them.
The alignment network is learnt on real samples by matching each video frame to its corresponding audio frame \cite{feng2023self}.
We call our method \ours (AV-HuBERT Aligned) and
show its depiction in Figure~\ref{fig:method}.

\mypar{Self-supervised features.}
We use AV-HuBERT to represent both the audio and visual content of a video.
AV-HuBERT is a Transformer network trained in a self-supervised way to predict iteratively refined centroids from masked inputs.
The features extracted by AV-HuBERT encode audio information as proved by its strong performance on tasks such as lip reading or noisy audio-visual speech recognition.
We extract the audio and visual representations independently:
to extract audio features we mask the visual input,
to extract visual features we mask the audio input.
For a video, we obtain audio features $\ab_i$ and visual features $\vb_i$ features for each time step $i$.
Both representations are 1024 dimensional and have a temporal resolution of 25 frames per second.

\mypar{Alignment network.}
To tell how well the audio and visual features match each other, we first independently L2 normalize the feature vectors and then feed them into a network $\Phi$.
This is implemented as a multi-layer perceptron (MLP) over the concatenated normalized audio and visual features:
\begin{align}
    \Phi_{ij} = \text{MLP}\left(\left[\ab_i; \vb_j\right]\right).
\end{align}
The MLP has four layers,
which progressively reduce the feature dimensionality,
with layers mapping from the AV-HuBERT feature size of 1024 to 512, 256, 128, and finally to a single output.
Each hidden layer includes Layer Normalization and ReLU activations.

\mypar{Loss function.}
To learn the alignment network $\Phi$ we maximize the probability of an audio frame $\ab_i$ to match the corresponding video frame $\vb_i$;
this probability is defined as:
\begin{align}
    p(\vb_i | \ab_i) = \frac{\exp\Phi_{ii}}{\sum_{k\in\mathcal{N}(i)}\exp\Phi_{ik}},
    \label{eq:loss}
\end{align}
where $\mathcal{N}(i)$ represents the temporal neighborhood around the frame $i$.
In our case $\mathcal{N}(i)$ contains the 30 neighboring frames around $i$.
We define the final loss as the negative probabilities averaged across the entire video:
\begin{align}
    \mathcal{L} = -\frac{1}{T} \sum_{i=1}^T \log p(\vb_i|\ab_j).
\end{align}
This loss is similar to the contrastive loss InfoNCE \cite{oord2018representation}, which was also used for deepfake detection \cite{feng2023self,oorloff2024avff}.

\mypar{Inference.}
Once $\Phi$ is learned, we can estimate the fakeness score as the negative of $\Phi_{ii}$ for each audio--video frame pair in a video;
aligned audio--video frame pairs should yield lower scores.
Then we compute an overall alignment score for the entire video by pooling the per-frame scores using the log-sum-exp function
(a smooth version of the max function).
 

\mypar{Supervised variant.}
To understand the impact of the silence bias on the standard supervised learning paradigm,
we design a supervised variant, \ourssup, that uses the 
same features and alignment network as \ours, but a classification loss.
In this setup 
we assume that apart from the real videos,
we also have access to fake videos in the training set, with corresponding labels $y$.
To obtain a per-video fakeness probability, we first pool the negated per-frame scores $\Phi_{ii}$ with the log-sum-exp function 
and then we apply the sigmoid function $\sigma$.
Finally, we optimize the binary cross-entropy (BCE) loss:
\begin{align}
    \mathcal{L}_{\text{sup}} = \text{BCE}\left(\sigma(\log\text{sum}\exp(-\Phi_{ii})), y\right).
\end{align}

\begin{table*}
\centering
\footnotesize
\tabcolsep 4.75pt
\begin{tabular}{%
    l ccc @{\hskip 0.2in}
    rr@{\hskip 0.01in}c
    rr@{\hskip 0.01in}c
    rr@{\hskip 0.01in}c
    rr@{\hskip 0.01in}c
}
\toprule
    &  &  &
    & \multicolumn{6}{c}{Metric: AUC}
    & \multicolumn{6}{c}{Metric: AP}
    \\
    \cmidrule(lr){5-10} \cmidrule(lr){11-14}
    &  &  &
    & \multicolumn{3}{c}{FakeAVCeleb} & \multicolumn{3}{c}{AV-Deepfake1M}
    & \multicolumn{3}{c}{FakeAVCeleb} & \multicolumn{3}{c}{AV-Deepfake1M}
    \\
    \cmidrule(lr){5-7} \cmidrule(lr){8-10} \cmidrule(lr){11-13} \cmidrule(lr){14-16}
    Method  & Modality & Train type & Train data
    & \multicolumn{1}{c}{Trim: \xmarkred} & \multicolumn{2}{c}{Trim: \checkmarkgreen}
    & \multicolumn{1}{c}{Trim: \xmarkred} & \multicolumn{2}{c}{Trim: \checkmarkgreen}
    & \multicolumn{1}{c}{Trim: \xmarkred} & \multicolumn{2}{c}{Trim: \checkmarkgreen}
    & \multicolumn{1}{c}{Trim: \xmarkred} & \multicolumn{2}{c}{Trim: \checkmarkgreen}
    \\
    \midrule
    \ourssup                    & AV & sup.   & FAVC      & 99.2 & 99.2 & \trg{} &  69.0 & 63.6 &  \trr{} & 100.0 & 100.0 & \trg{} &  84.1 & 82.2 & \trre{} \\
    \ourssup                    & AV & sup.   & AV1M      & 77.5 & 70.8 & \trr{} & 100.0 & 83.1 &  \trr{} &  99.4 &  99.1 & \trg{} & 100.0 & 94.4 & \trr{} \\
    \midrule 
    AVAD \cite{feng2023self}    & AV & unsup. & LRS       & 84.5 & 84.7 & \trg{} &  54.3 & 54.3 &  \trg{} &  99.5 &  99.5 & \trg{} &  76.3 & 76.3 & \trg{} \\
    SpeechForensics \cite{liang24speechforensics}  
                                & AV & unsup. & VoxCeleb2 & 98.8 & 98.8 & \trg{} &  68.8 & 68.2 &  \trg{} & 100.0 & 100.0 & \trg{} &  83.7 & 83.5 & \trg{} \\
    \ours                       & AV & unsup. & VoxCeleb2 & 94.6 & 94.6 & \trg{} &  85.9 & 83.5 & \trre{} &  99.8 &  99.8 & \trg{} &  94.3 & 93.5 & \trg{} \\
\bottomrule
\end{tabular}
\caption{%
    Comparison of \ours and \ourssup  on FakeAVCeleb (FAVC) and AV-Deepfake1M (AV1M) original and trimmed datasets.
    \ours is not impacted by the presence of leading silence in the original datasets, showing similar performance on the trimmed versions.
    In contrast, the AUC performance of \ourssup degrades by $16.9\%$ when tested on the trimmed variant of AV1M.
}
\label{tab:fakeavceleb_av1m_trimmed}
\end{table*}
\newlength{\widthpic} 
\setlength{\widthpic}{0.3\textwidth} 
\begin{figure*}
\begin{tabular}{ccc}
      \includegraphics[width = \widthpic]{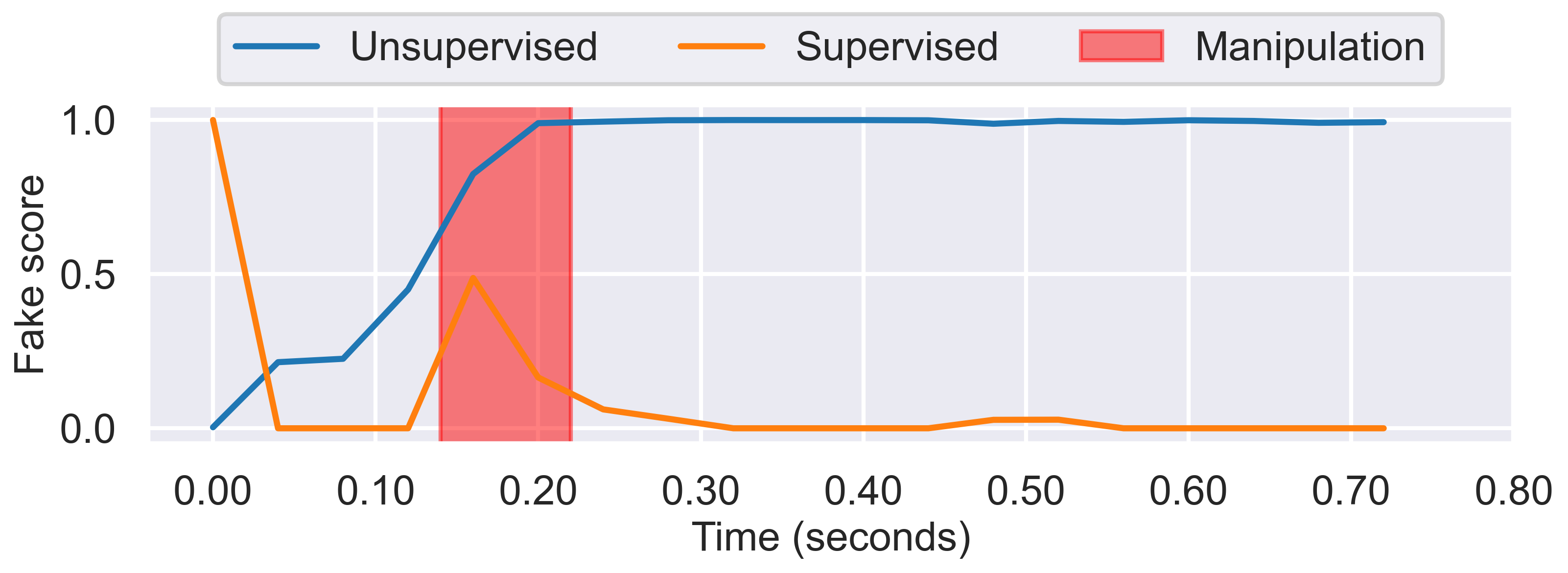} & 
      \includegraphics[width = \widthpic]{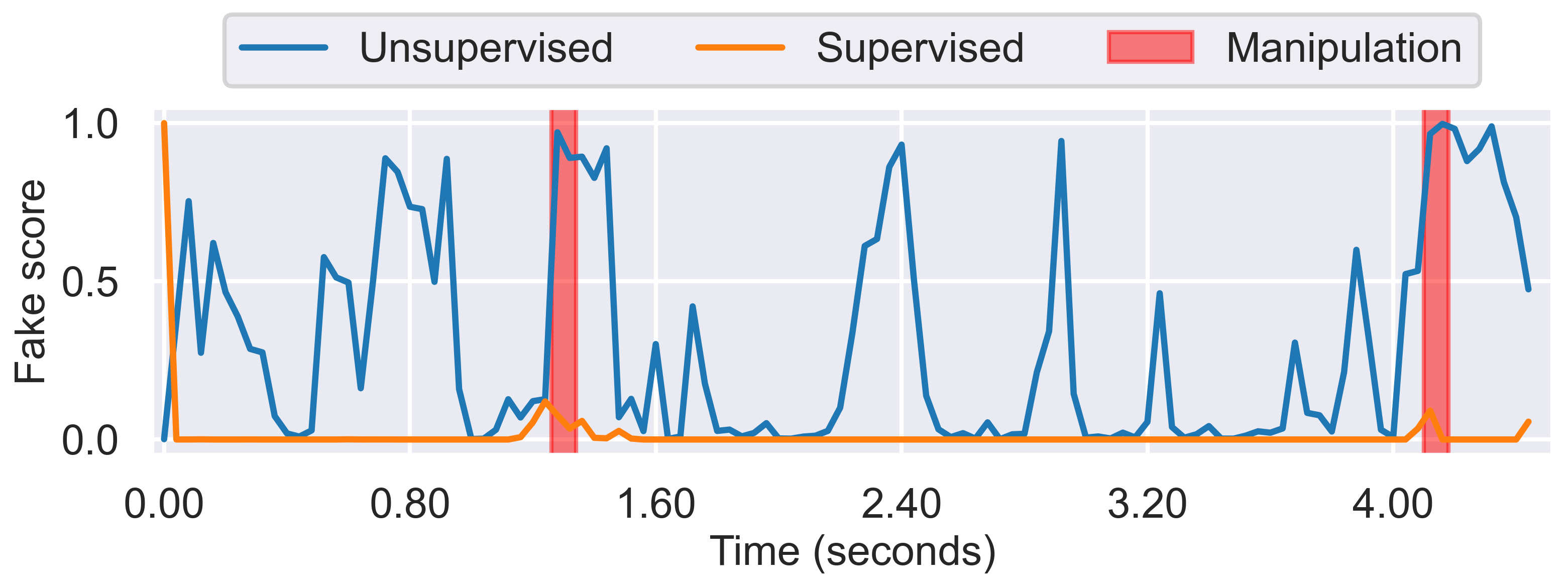} &
      \includegraphics[width = \widthpic]{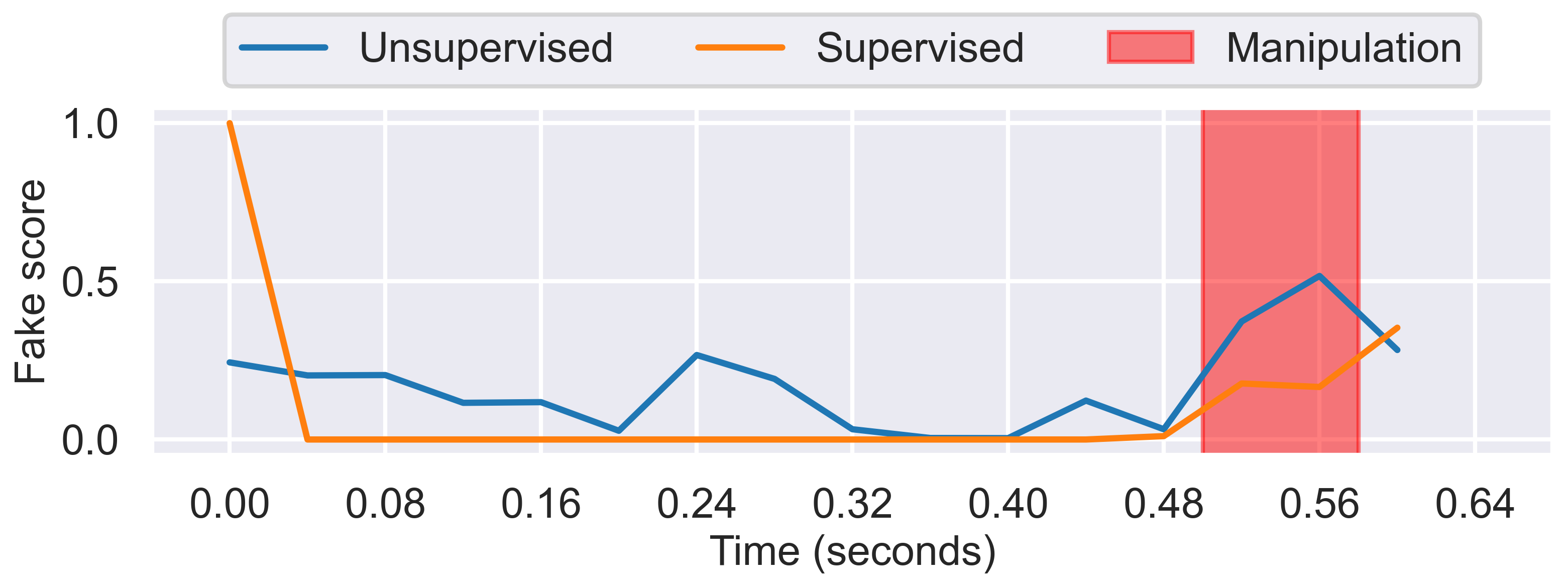} \\
      \includegraphics[width = \widthpic]{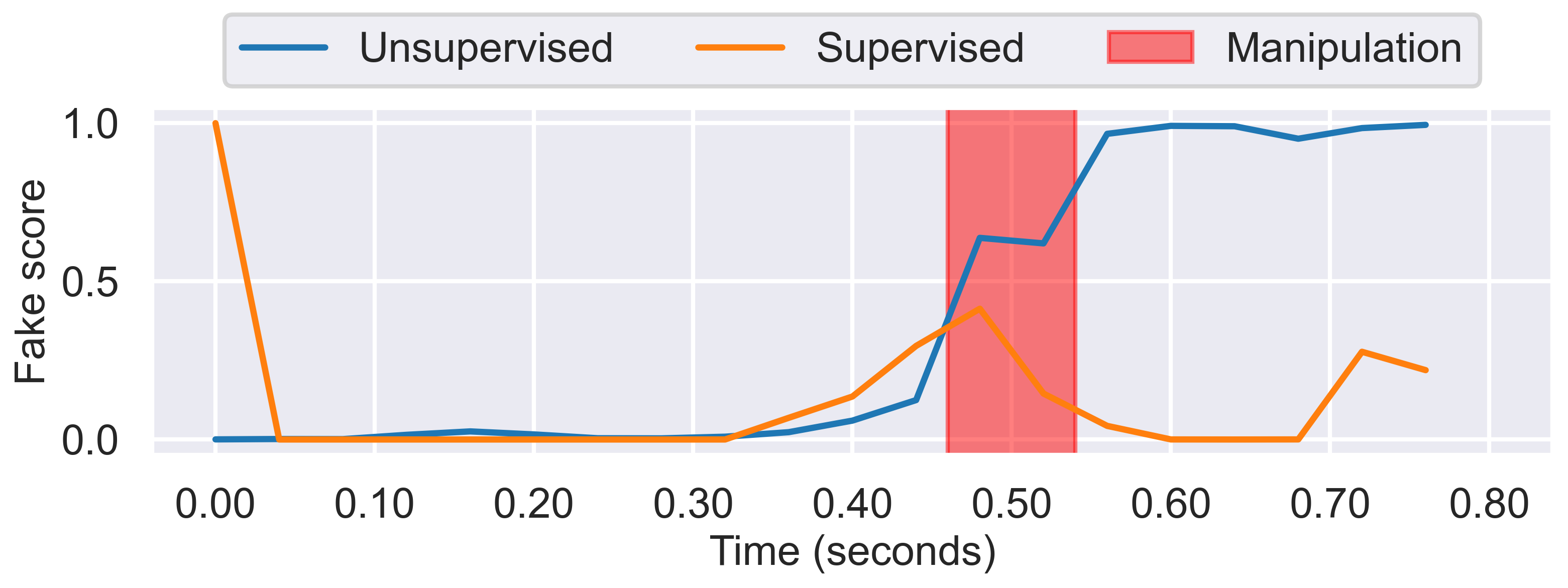}&
      \includegraphics[width = \widthpic]{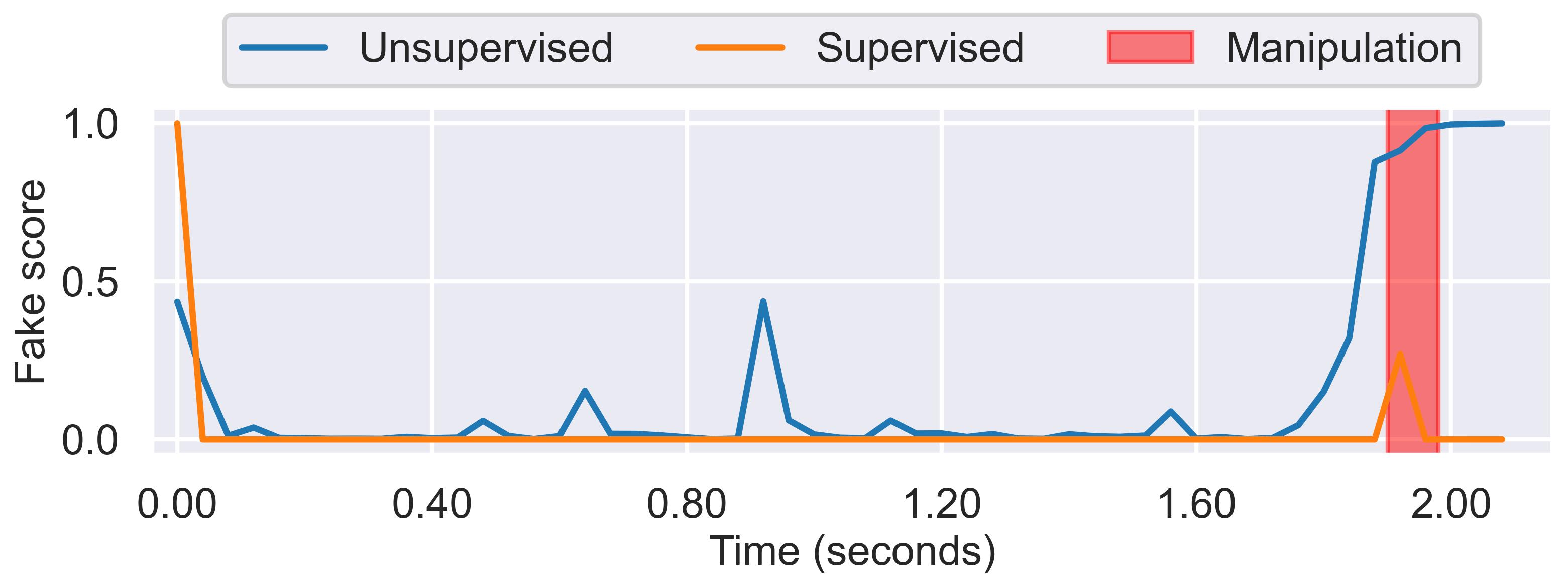} &
      \includegraphics[width = \widthpic]{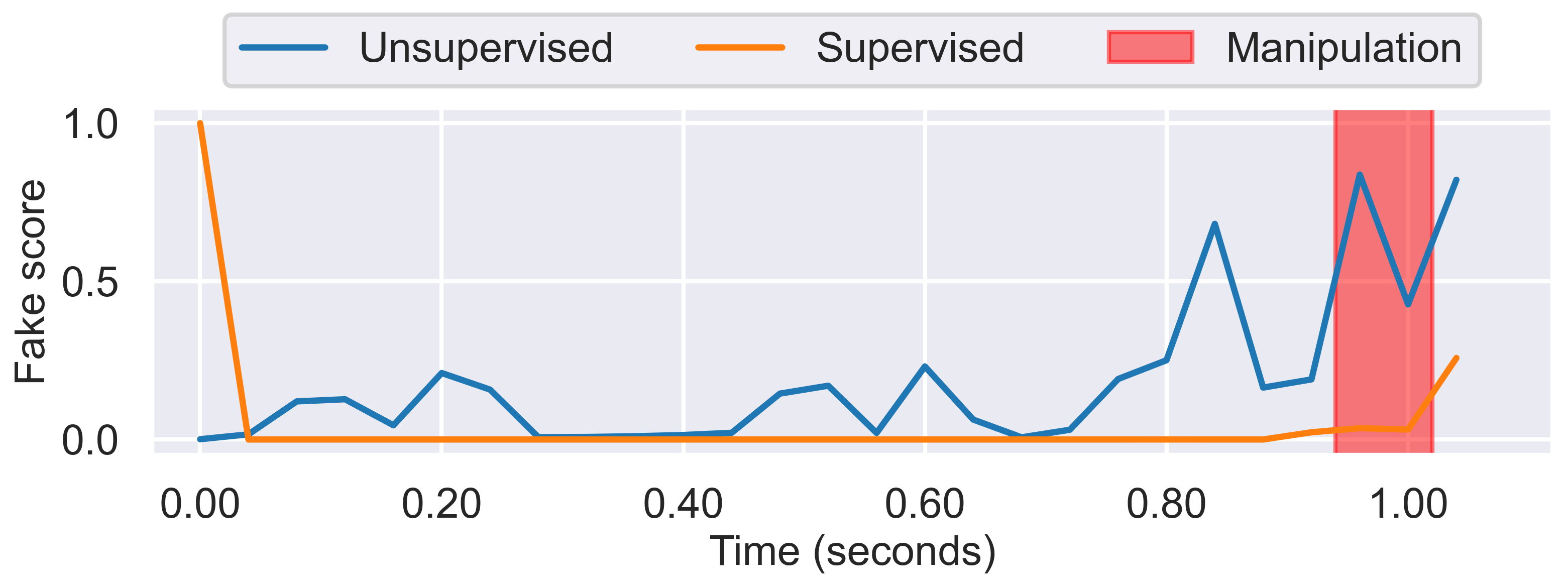} \\
      \includegraphics[width = \widthpic]{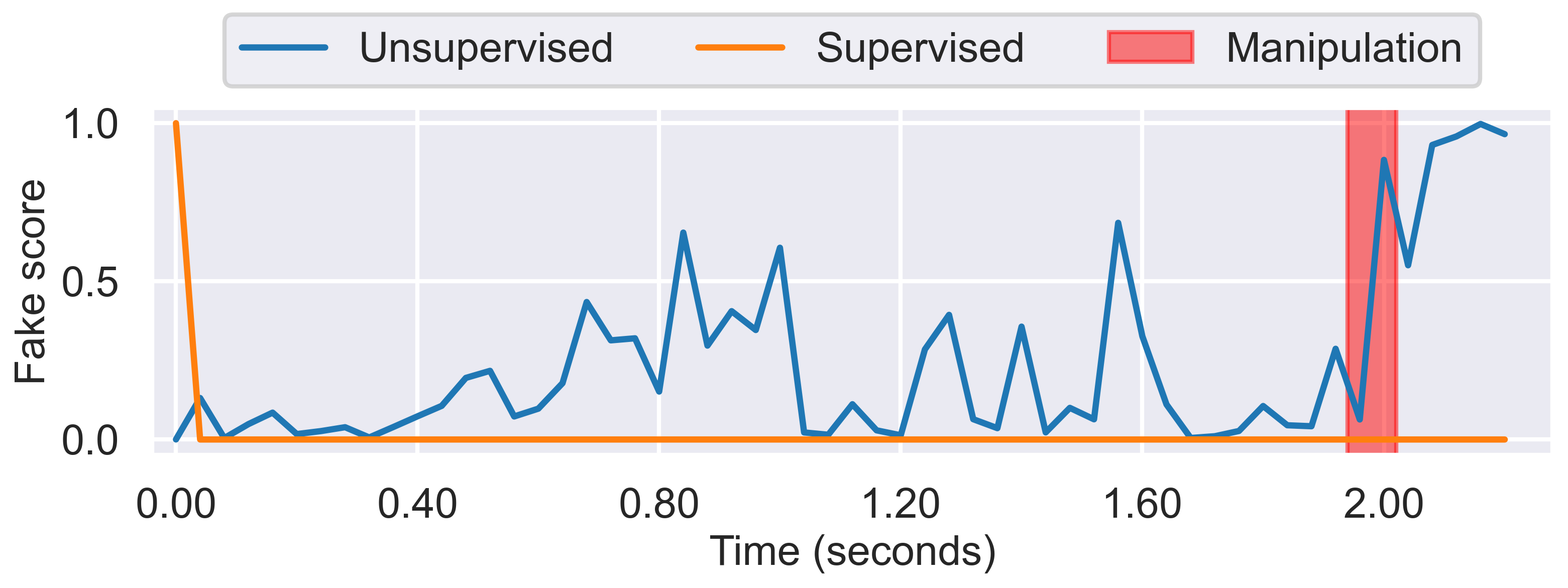} &
      \includegraphics[width = \widthpic]{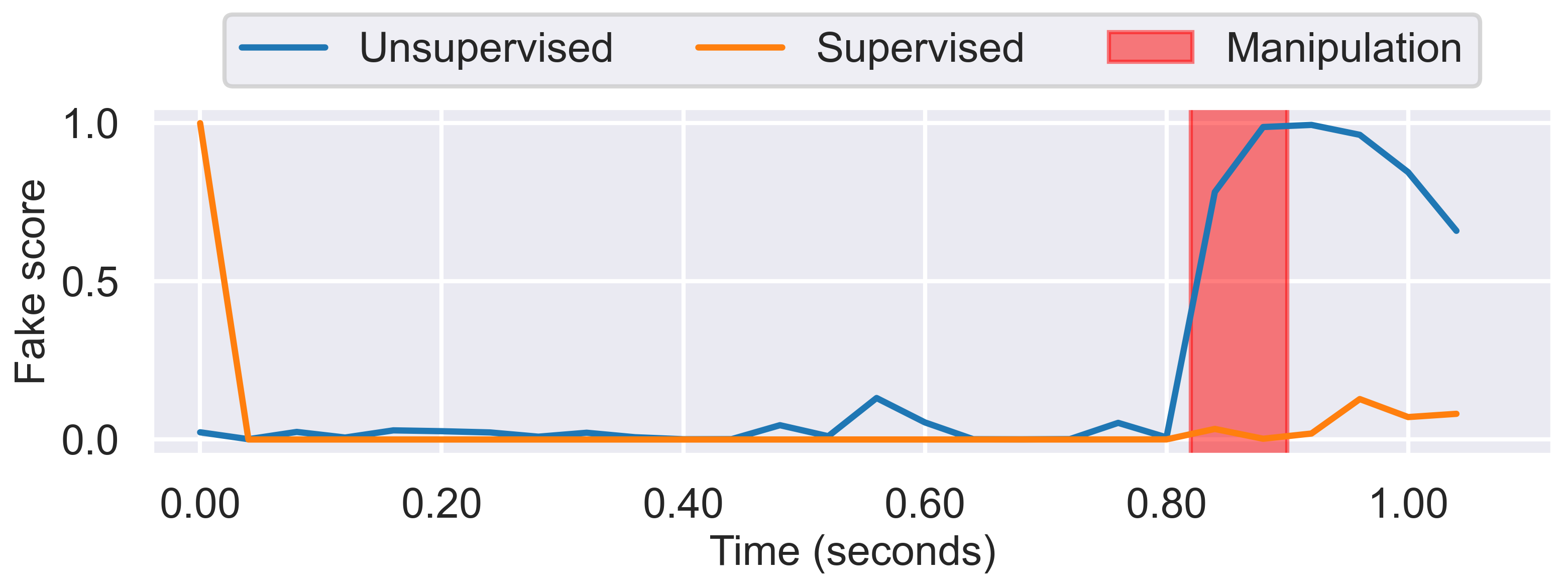} &
      \includegraphics[width = \widthpic]{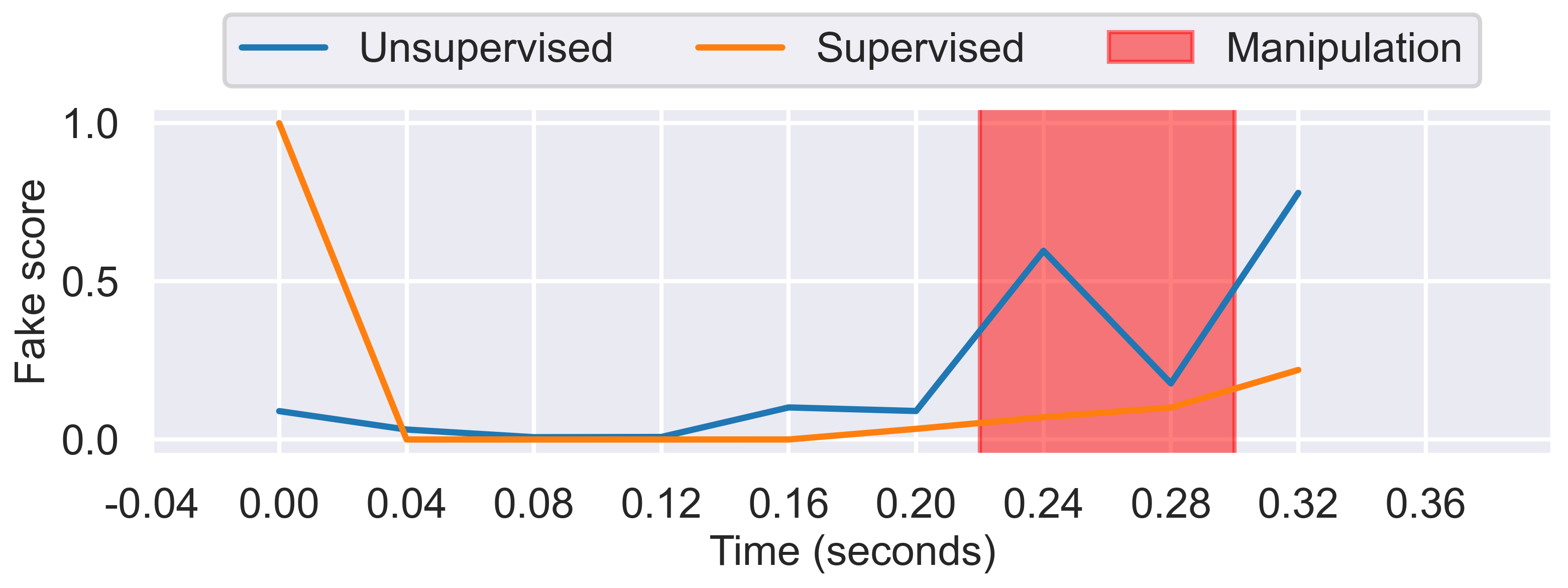} \\
\end{tabular}
\caption{%
Per frame fakeness probabilities for \ours and \ourssup on AV-Deepfake1M.
\ourssup always marks the first frame---corresponding usually to the leading silence---as fake, thus confirming that it uses the bias to distinguish between real and fake videos.
\ours is not affected by the presence of the leading silence.
The fakeness probabilities for \ours can be interpreted as misalignment probabilities, which is why they are higher during or after the manipulated region. 
}
\label{fig:per_frame_scores}
\end{figure*}

\subsection{Experimental setup}

\mypar{Datasets and metrics.}
We conducted our experiments on the two datasets introduced in Sec. \ref{subsec:datasets}: FakeAVCeleb and AV-Deepfake1M.
For FakeAVCeleb, we use 70\% of the dataset for training and validation, and the rest of 30\% for testing. 
For AV-Deepfake1M, we evaluate on 10k samples from the original validation set,
but for the best performing models we also report results on the official withheld test set.
To train \ours, we use 50k real samples from the training set of AV-Deepfake1M (45k samples for training and 5k for validation). 
Since AV-Deepfake1M is based on VoxCeleb2, these samples originate from VoxCeleb2.
To train the supervised variant, \ourssup, on AV-Deepfake1M, we select the same number of samples (45k for training and 5k for validation), but this time coming from both the real and fake classes.
For both datasets, we evaluate video-level detection.
Following prior work, 
we report results in terms of the area under the receiver operator characteristic curve (AUC) 
and average precision (AP), with the fake class being the positive class.

\mypar{Implementation details.}
For \ours we use a learning rate scheduler with a patience of 5 epochs and a factor of 0.1, with a starting learning rate of $10^{-5}$. The training is stopped if there is no loss improvement on the validation set for 10 consecutive epochs. \ourssup is learned using an Adam optimizer with a learning rate of $10^{-3}$.
For the AV-HuBERT feature extractor
we use the checkpoint pretrained on LRS3 and VoxCeleb2 and finetuned on LRS3 for visual speech recognition.



\subsection{Experimental results}


We train our method, \ours, as well as its supervised variant, \ourssup, on the two datasets, FakeAVCeleb and AV-Deepfake1M, and evaluate on either of them (that is, both in-domain and out-of-domain).
For each evaluation, in addition to the original set, we also consider the trimmed version of the validation dataset (trim: \checkmarkgreen).
The trimming is performed as described in Sec.~\ref{subsec:impact-on-prior-work}.
We always use the untrimmed, original data for training.
Results are shown in Table \ref{tab:fakeavceleb_av1m_trimmed}.

\mypar{Impact of leading silence bias.}
We observe that \ours is robust to the leading silence bias:
trimming the silence has no effect on FakeAVCeleb and only a slight effect on AV-Deepfake1M ($2.4\%$ AUC).
On the other hand, the performance of the supervised variant, \ourssup, decreases considerably when testing on AV-Deepfake1M
(by $16.9\%$ or $5.4\%$ AUC, depending on the train set).
On FakeAVCeleb removing the leading silence has no impact for \ourssup, when training also on FakeAVCeleb.
This may happen because FakeAVCeleb has full manipulations and there is other useful information in the video.
Instead, when training on the locally-manipulated AV-Deepfake1M, the difference becomes significant ($6.7\%$ AUC).

\mypar{Comparison with other unsupervised methods.}
We compare our results with those of AVAD \cite{feng2023self} and SpeechForensics \cite{liang24speechforensics},
which are also usupervised methods, trained on real data only.
Similar to \ours they are not impacted by the spurious leading silence, showing nearly identical results for trimmed and untrimmed datasets.
However, their overall performance is considerably worse than ours on the AV-Deepfake1M dataset. 

\mypar{Score visualization.} 
Figure \ref{fig:per_frame_scores} shows the per frame scores obtained by the two methods, \ours and \ourssup,
together with the groundtruth manipulated interval.
For the \ours method, the scores represent the misalignment probability between the audio and visual streams at each time frame;
for the \ourssup method, the scores represent the probability of an audio-video frame of being fake.
We see that \ourssup always predicts the first frame as being fake,
confirming once again that it has learned to associate the spurious leading silence with a video being fake.
When detected, the actual fake region is given a considerable lower fakeness probability than that assigned to the leading silence.
On the other hand, \ours is not affected by the presence of leading silence. 
In this case the manipulated areas have higher misalignment scores,
a frame-level evaluation yielding 77.7\% AUC.
We do notice that there are other regions in the video that get a high misalignment score.
This happens because the separation and re-composition of the audio and video streams can introduce desynchronizations between the two even in areas that were not intentionally manipulated.

\subsection{Comparison on the official AV-Deepfake1M test}

\begin{table}
\centering
\footnotesize
\begin{tabularx}{\linewidth}{Xccc}
\toprule
     Methods & Modality & AUC \\
    \hline
    \textit{Segment-level methods} \\
    Meso4 \cite{Afchar_2018} & V & 54.53 \\
    MesoInception4 \cite{Afchar_2018} & V & 57.16 \\
    MDS \cite{chugh2020not} & AV & \textcolor{my_red}{56.57} \\
    MARLIN \cite{cai2023marlinmaskedautoencoderfacial} & V  & 58.03 \\
    \hline
    \textit{Frame-level methods} \\
    Meso4 \cite{Afchar_2018} & V & 63.05 \\
    MesoInception4 \cite{Afchar_2018} & V & 64.04 \\
    Xception \cite{8099678} & V & 68.68 \\
    EfficientViT \cite{coccomini2022iciap} & V  & 65.51 \\
    \hline
    \textit{Ours} \\
    \ours & AV &  \textbf{85.24} \\
    \ourssup & AV & \textcolor{my_red}{99.90} \\
    Silence classifier & A & \textcolor{my_red}{98.44} \\
\bottomrule
\end{tabularx}
\caption{Results on the official AV-Deepfake1M test set.
Except the methods that have access to the leading silence bias (shown in red),
\ours performs best with 85.24$\%$ AUC.
Having access to the leading silence bias,
the silence classifier and \ourssup show almost perfect performance of over $98\%$ AUC.}
\label{tab:results_test}
\end{table}

To further verify our conclusions,
we evaluate our approach on the official test set of the AV-Deepfake1M dataset for which the labels are witheld.
The main difference between the validation set, used in the previous section, and the test set are the subject identities:
the test subjects are unseen at train time.
We obtain the results by submitting our prediction to the official competition server.
Our results are compared against those of other methods based on the values reported in the AV-Deepfake1M paper \cite{cai2024avdeepfake1mlargescalellmdrivenaudiovisual}. 

The results shown in Table \ref{tab:results_test} are in line to those obtained on the validation split (Table~\ref{tab:fakeavceleb_av1m_trimmed}).
Compared to the rest of the methods, \ours has the highest performance (85.24$\%$ AUC) even if has seen only real data at training.
The silence classifier described in Sec.~\ref{sec:silence-bias} obtains again a performance of over $98\%$ AUC, indicating that the test set suffers from the same spurious feature as the training and validation splits.
Other methods that are trained at video level do not have access to the silence bias,
and hence show only modest performance.
Instead, we expect audio-based methods to use the bias and return over optimistic results.
This is indeed the case for the supervised variant, \ourssup, which returns an AUC of $99.90\%$ on the official test set.

\subsection{Further analysis}

\mypar{Ablation of \ours components.} 
Here we analyze the impact of various design choices on the performance of the \ours method.
Specifically, we investigate the following:
1) removing feature normalization;
2) reducing the training set size from 45,000 to 8,782 samples (the real ones from the \ourssup train set);
3) using mean score pooling instead of log-sum-exp;
4) extracting features with the AV-HuBERT checkpoint pretrained on LRS3 only  instead of LRS3 and VoxCeleb2;
5) using a linear layer instead of MLP.
Results are shown in Table \ref{tab:unsupervised_ablations}.
All ablations lead to a decrease in performance with the exception of mean score pooling for FakeAVCeleb and AVLips.
This is expected, however, since fake videos in FakeAVCeleb and AVLips are manipulated at every frame,
while AV-Deepfake1M has only locally-manipulated fake videos. 

\begin{table}
    \centering
    \footnotesize
    \newcommand{\ii}[1]{\textcolor{gray}{#1}}
    \begin{tabularx}{\linewidth}{r X rrr}
        \toprule
        & Configuration & FAVC & AV1M & AVLips \\
        \midrule
        \ii{ } & \ours                          &   \underline{94.6} & \bf 85.9 & \underline{86.3} \\
        \ii{1} & \dots\ Feature norm: \xmarkred &   92.4 &     \underline{84.8} & 86.1 \\
        \ii{2} & \dots\ Train size: 8.7k        &  89.1 &     82.5 & 83.0 \\
        \ii{3} & \dots\ Score pooling: mean     & \bf 97.4 &     76.8 & \bf 88.8 \\
        \ii{4} & \dots\ AV-HuBERT: LRS3         &  84.8 &     60.3 &  80.1 \\
        \ii{5} & \dots\ Alignment net: Linear   &  31.5 &    51.9 & 43.4 \\
        \bottomrule
    \end{tabularx}
    \caption{Ablation of the main components of \ours in terms of AUC.
    Mean score pooling helps when videos are fully generated (FAVC, AVLips).
    Other ablations degrade the performance.
    } 
    \label{tab:unsupervised_ablations}
\end{table}

\mypar{Impact of VoxCeleb2 pretraining.}
Both FakeAVCeleb and AV1M datasets use VoxCeleb2 as their source dataset.
Can there be a data leakage from AV-HuBERT VoxCeleb2 features to these datasets?
We evaluate on a dataset that is not using VoxCeleb2, AVLips \cite{liu2024lips}.
We observe that VoxCeleb2 features perform better even on this dataset,
suggesting that these features are simply stronger than LRS3 features.
This is may be explained by the size of VoxCeleb data (about four times larger than LRS3)
and is in line with the results on other downstream tasks (e.g., lip reading \cite{shi2022learning}).

\mypar{Analysis of \ourssup architecture.}
We investigate how a simpler alignment network impacts the supervised model.
To this end, we train a linear layer on top of the AV-HuBERT representations instead of the MLP network.
In Table~\ref{tab:supervised_architecture}, 
we observe that unlike the unsupervised case where a linear layer was too weak to learn,
here a linear layer performs as well or even better than the MLP.
The largest difference between the MLP and then linear layer is observed when training on FakeAVCeleb and testing on AV-Deepfake1M.
With a linear layer, the performance drop from untrimmed to trimmed AV-Deepafake1M is $19.3\%$ (from $85.9\%$ to $66.6\%$ AUC),
while with an MLP, the drop is $5.4\%$ (from $69.0\%$ to $63.6\%$ AUC). 

\begin{table}[htb!]
\centering
 \footnotesize
\tabcolsep 3pt
\begin{tabular}{lc rr rr}
\toprule
             & & \multicolumn{2}{c}{FakeAVCeleb}  & \multicolumn{2}{c}{AV-Deepfake1M}\\
          \cmidrule(lr){3-4}
           \cmidrule(lr){5-6}
Architecture   & Train data  & Trim: \xmarkred & Trim: \checkmarkgreen & Trim: \xmarkred  & Trim: \checkmarkgreen \\
\midrule
Linear  & FakeAVCeleb  & 99.6 & 99.5 & 85.9 & 66.6 \\
MLP  & FakeAVCeleb  & 99.2 & 99.2 & 69.0 & 63.6 \\
\hline
Linear  & AV-Deepfake1M  & 76.2 & 69.7 & 99.9 & 83.2 \\
MLP  & AV-Deepfake1M  & 77.5 & 70.8 & 100.0 & 83.1 \\
\bottomrule
\end{tabular}
\caption{%
    Architecture analysis of \ourssup.
    In most cases, the linear and MLP architectures perform similarly.
} 
\label{tab:supervised_architecture}
\end{table}

\section{Discussion}

We discuss how our conclusions fit into the broader scope of deepfake detection.

\mypar{A different evaluation paradigm.}
We observed near-perfect performance on datasets such as FakeAVCeleb and AV-Deepfake1M.
The reasons are that the same generative models were used in both train and test,
as well as the unintentional leakage of spurious features.
The latter also affects cross-dataset performance, \eg, training on FakeAVCeleb and testing on AV-Deepfake1M still yields strong performance.
In a realistic scenario, however, we are not given access to the generators,
nor to the pre-processing or post-processing steps,
which introduce unintentional alterations. 
Each deepfake released in the wild may be created through a different set of tools, unknown apriori. 
As such, we believe that another way of gauging the progress of deepfake detection 
is by refraining the training to real data only.

\mypar{Circumventing shortcuts through alignment.}
Modeling real data is a way to avoid shortcuts in the data.
A different direction is taken in the very recent work of Rajan \etal \cite{rajan2024}.
The idea is to generate fake samples by reconstructring real samples through a generator of choice.
This approach ensures that fakes and reals are aligned,
avoiding spurious features. 
On the other hand, this forces the model to focus on the fingerprint of the generator,
hindering generalizability.
Moreover, by equating fake to a fingerprint,
we become susceptible to laundering \cite{mandeli2024},
lose the ability to localise \cite{mareen2024,smeu2024} and distinguish benign fakes (\eg, superresolution images \cite{yan2024}).
We believe that both directions (alignment and real data)
are 
complementary perspectives that should be considered to tackle the multi-faceted problem of deepfake detection.

\section{Conclusions}
In this paper we exposed a previously unknown bias in two widely adopted audio-video deepfake detection datasets---a leading silence in fake videos.
We showed that models exposed to this bias during training are prone to rely on it when deciding the authenticity of a video, thus displaying overly optimistic results.
As an alternative, we propose to shift the learning paradigm towards unsupervised learning on real data only.
Specifically, we 
find that
self-supervised audio-video representations coupled with an alignment network trained on real videos 
produce more robust and consistent results.
Our work raises awareness regarding dataset design and evaluation of deepfake detection.

\section*{Acknowledgments}
This work was supported in part by the EU Horizon projects AI4TRUST (No. 101070190) and ELIAS (No. 101120237),
and by CNCS-UEFISCDI (PN-IV-P7-7.1-PTE-2024-0600).

{
    \small
    \bibliographystyle{ieeenat_fullname}
    \bibliography{main}
}


\end{document}